\documentclass[10pt,twocolumn,letterpaper]{article}

\usepackage[pagenumbers]{cvpr} 

\definecolor{cvprblue}{rgb}{0.21,0.49,0.74}
\usepackage[pagebackref,breaklinks,colorlinks,allcolors=cvprblue]{hyperref}
\usepackage{enumitem}
\usepackage{floatrow}
\newfloatcommand{capbtabbox}{table}[][\FBwidth]
\usepackage{multirow} 
\usepackage{multicol}
\usepackage{wrapfig}
\usepackage{amsmath}

\title{EMatch: A Unified Framework for Event-based Optical Flow and \\ Stereo Matching}
\author{Pengjie Zhang, Lin Zhu, Xiao Wang, Lizhi Wang, Wanxuan Lu, Hua Huang\\ 
Beijing Institute of Technology, Beijing Normal University\\
}

\begin{document}
\maketitle
\begin{abstract}
Event cameras have shown promise in vision applications like optical flow estimation and stereo matching, with many specialized architectures leveraging the asynchronous and sparse nature of event data. However, existing works only focus event data within the confines of task-specific domains, overlooking how tasks across the temporal and spatial domains can reinforce each other. In this paper, we reformulate event-based flow estimation and stereo matching as a unified dense correspondence matching problem, enabling us to solve both tasks within a single model by directly matching features in a shared representation space. Specifically, our method utilizes a Temporal Recurrent Network to aggregate event features across temporal or spatial domains, and a Spatial Contextual Attention to enhance knowledge transfer across event flows via temporal or spatial interactions. By utilizing a shared feature similarities module that integrates knowledge from event streams via temporal or spatial interactions, our network performs optical flow estimation from temporal event segment inputs and stereo matching from spatial event segment inputs simultaneously. We demonstrate that our unified model inherently supports multi-task fusion and cross-task transfer. Without the need for retraining for specific task, our model can effectively handle both optical flow and stereo estimation, achieving state-of-the-art performance on both tasks. Our code will be released upon acceptance.
\end{abstract}    
\section{Introduction}
Event cameras are novel neuromorphic vision sensors that detect intensity changes at each pixel, offering advantages like low latency, high temporal resolution, and high dynamic range \cite{128128, eSurvey}. In the context of 3D scene understanding, event-based vision tasks can be broadly divided into two categories: temporal tasks and spatial tasks. Temporal tasks aim to understand motion within 3D scenes, such as optical flow \cite{evflownet, eraft, TMA} and object tracking \cite{jiang2020object}, while spatial tasks focus on comprehending the structure of 3D objects, such as stereo matching \cite{DDES, DTC, ConcentrationNet} and depth estimation \cite{EMVS}.

To date, most event-based research has concentrated on only one of these aspects, without seeking to unify temporal and spatial perception. We focus on optical flow estimation and stereo matching, where prior works have developed specialized frameworks for efficiently extracting task-specific features. In optical flow estimation, most frameworks derive temporal motion features either directly from accumulated events \cite{evflownet, STEFlowNet, IDNet} or by constructing an event-based cost volume \cite{eraft, ADMFlow, EFlowFormer, TMA}. For stereo matching, many frameworks \cite{DDES, ConcentrationNet, DTC, ETAM-TSCLM} employ a classic approach to compute spatial matching costs from events.

\begin{figure*}
\begin{center}
\includegraphics[width=\linewidth]{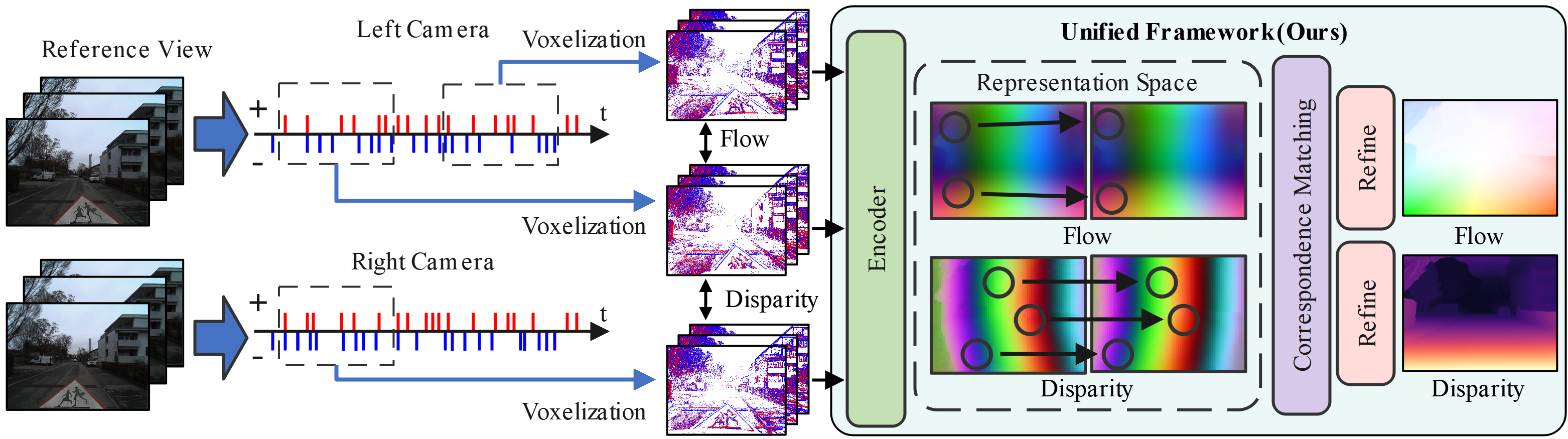}
\end{center}
\vspace{-4mm}
\caption{Overview of our unified framework. Previous works considered optical flow estimation and stereo matching as two separate tasks and designed many frameworks in their respective pipeline. We reformulate these two tasks as a dense correspondence matching problem and design a novel unified framework with a shared representation space.}
\label{figure1}
\end{figure*}

However, existing works are confined to task-specific domains, overlooking the potential for cross-domain reinforcement between temporal and spatial tasks. Both flow estimation and stereo matching, in fact, can be treated as dense correspondence matching problems \cite{unify, Flow2stereo, HD3, DPCTF}. As shown in Fig.~\ref{figure1}, flow represents the temporal difference between event segments captured at different times, while disparity represents the spatial difference between event segments from different viewpoints. Building on this insight, we propose to estimate flow and disparity from events through dense correspondence matching, unifying both tasks within a shared representation space.

In this paper, we reformulate event-based flow estimation and stereo matching as a unified dense correspondence matching problem. Specifically, we introduce a Temporal Recurrent Network (TRN) and Spatial Contextual Attention (SCA) to map the initial event streams into a shared representation space. First, since events are triggered asynchronously, we accumulate them to obtain an event voxel representation \cite{zhu2019unsupervised, eraft, TMA}. To fully utilize the temporal information in the event voxel, we iteratively extract temporal features with the TRN. Second, as events are only triggered at pixels with intensity variations, event data is unevenly distributed in space. To address this, we assign values to each pixel using contextual information through the SCA. By aggregating event features temporally and spatially, we generate a unified feature map for dense correspondence matching.

Building on TRN and SCA, we propose EMatch, a novel event-based model that serves as a unified framework for optical flow estimation and stereo matching. Unlike previous task-specific models, EMatch can be used for optical flow estimation, stereo matching, or even both tasks simultaneously. Moreover, its unified architecture facilitates cross-task transfer, making it adaptable to different tasks without requiring extensive modifications. Experiments demonstrate that EMatch achieves state-of-the-art performance on the DSEC benchmark for both optical flow estimation and stereo matching, while also excelling in multi-task fusion and cross-task transfer.

Our main contributions can be summarized as follows:

(1) We propose EMatch, a novel event-based framework that unifies optical flow estimation and stereo matching within a shared representation space using dense correspondence matching. Our framework bridges the gap between temporal and spatial perception, enabling the simultaneous handling of motion and stereo estimation.

(2) We introduce two key modules Temporal Recurrent Network (TRN) and Spatial Contextual Attention (SCA). Together, TRN and SCA generate a unified feature map for dense correspondence matching by aggregating event features temporally and spatially.

(3) Our unified model inherently supports multi-task fusion and cross-task transfer, achieving state-of-the-art performance in both optical flow estimation and stereo matching within a single unified architecture.

\section{Related Works}

\noindent \textbf{Optical Flow Estimation.} Traditionally, optical flow estimation relies on variational approaches \cite{LK}, where it is commonly tackled as an energy minimization problem. With the rise of deep learning, many works pioneered the application of neural networks \cite{flownet, pwcnet, raft}, and introduced attention mechanism to them \cite{gma, flowformer}. These deep-learning methods all establish a correlation layer and a regression head. Instead, GMFlow \cite{gmflow} reformulates optical flow estimation as a global matching problem to directly find dense correspondence between pixels, which provides inspiration for us to design an event-based matching framework.

For event-based optical flow, numerous studies have investigated strategies for effectively leveraging event data \cite{LK_E, barranco2014contour, zhu2017event, akolkar2020real, CM, Secrets}. Benosman et al.\cite{LK_E} first proposed a gradient-based method to estimate flow. Gallego et al. \cite{CM, Secrets} proposed a framework with the principle of contrast maximization. With the rise of deep-learning, many works introduced neural networks to learn priors from event data, and they adopted two different solutions. Firstly, some works estimate flow directly from one consecutive event stream \cite{evflownet, zhu2019unsupervised, STEFlowNet, IDNet}. Secondly, some works adopted a correlation-regression framework to estimate flow from two consecutive event streams \cite{eraft, EFlowFormer, ADMFlow, TMA}. In our paper, we adopt the second technical solution but introduce a novel framework to adpot dense correspondence matching between event streams.

\noindent \textbf{Stereo Matching.} The main objective of stereo matching is to find the disparity value based on rectified image pairs, which can be used to estimate depth. Traditionally, classical algorithms all adopted a regression pipeline \cite{SGM, PatchMatch, ADCensus}. Leveraging deep learning,  end-to-end methods achieved significant advancements in matching and optimization while maintaining traditional pipeline \cite{GCNet, PSMNet, GANet}. Unlike them, RAFT-Stereo \cite{raftstereo} introduced another iterative regression framework. GMStereo \cite{unify} simultaneously performs optical flow, disparity, and depth estimation on images through feature similarity computation, offering a simpler and more efficient approach.

For event-based stereo matching, early methods attempt to predict disparity by determining corresponding events with hand-crafted descriptors \cite{rogister2011asynchronous, camunas2014use, kogler2011event, piatkowska2013asynchronous, zhu2018realtime}. With the application of deep learning, neural networks are used to exploit the potential of event data \cite{DDES, DTC, ConcentrationNet, ETAM-TSCLM, TESNet, EIStereo}. DDES \cite{DDES} was the first learning-based model with a learnable representation for event data. DTC \cite{DTC} encoded high dimensional spatial-temporal features using proposed discrete time convolution. Se-cff \cite{ConcentrationNet} concentrated event stacks with multiple density events by an attention-based concentration network. Their designs are ingenious but intricate. Instead, our framework is more streamlined, with the only objective of finding correspondence between pixels. 

\noindent \textbf{Unified Methods.} Optical flow estimation and stereo matching are often jointly tackled for their geometrical consistency \cite{Geonet, DFNet, SENSE, CC, Unos}. However, they often deployed task-specific networks for each task, which increased the complexity of models. Therefore, some works attempted to design a model that is applicable to both optical flow and stereo matching with a shared network \cite{unify, Flow2stereo, HD3, DPCTF}. They considered both optical flow and disparity as dense correspondence between pixels, which inspired us to design a unified model for event camera.

Currently, few event-based works tackled optical flow estimation and stereo matching simultaneously. EVFlownet \cite{zhu2019unsupervised} proposed a framework applicable to flow or depth, but it ignored the association between them and only trained them separately. TESNet \cite{TESNet} introduced flow to guide training of stereo matching, but it still adopted extra non-associated network to estimate flow and the final output was only disparity. In our paper, we design a unified framework to adopt dense correspondence matching for flow and stereo, capable of estimating both flow and disparity simultaneously with shared architecture and parameters.

\section{Unified Model for Event-based Optical Flow and Stereo Matching}

As mentioned earlier, we need to estimate optical flow and disparity from event streams using correspondence matching within a shared representation space. In the following, we will firstly clarify how correspondence matching unifies the estimation of flow and disparity from event streams. Then, we will provide the details of our unified model, EMatch, mainly about how to use TRN and SCA to generate a high-dimensional feature map for dense matching. Additionally, there are some implementation details including matching algorithm and optimization strategies. 

\subsection{Dense Correspondence Matching for Events} \label{Sec3.1}

\begin{figure}
\begin{center}
\includegraphics[width=0.95\linewidth]{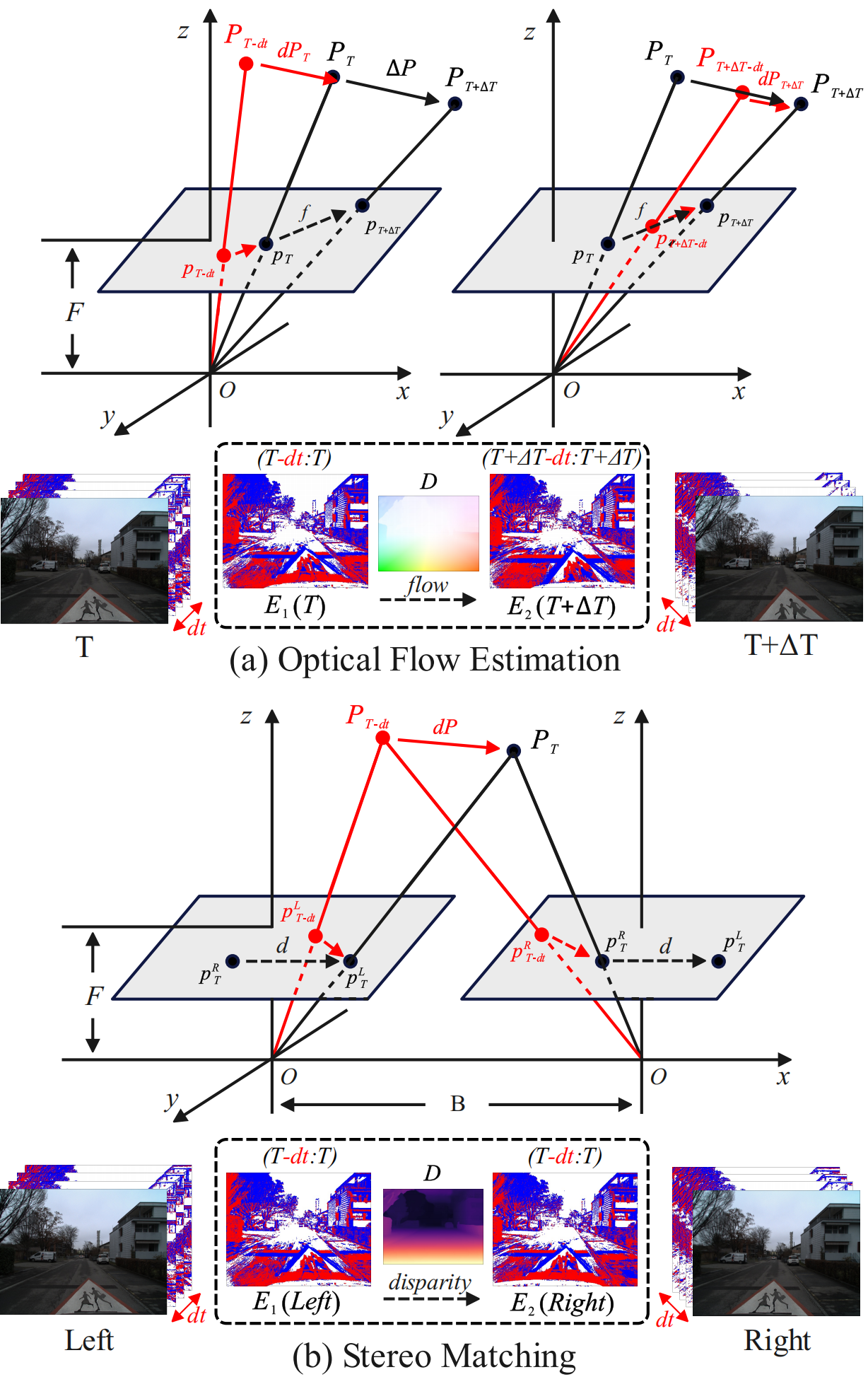}
\end{center}
\vspace{-6mm}
\caption{{Illustration of the similarity between event-based optical flow estimation and stereo matching by correspondence matching.} We accumulate events during a sampling time $dt$ from different times ($T, T+\Delta T$) or different viewpoints ($Left, Right$) to get reference and target event streams $E_{1}, E_{2}$. By correspondence matching, we can get the displacement $D$ (i.e. flow or disparity) between them. Note that the event streams needs to be transformed into a high-dimensional domain through feature extraction $\mathcal{F}(\cdot)$.}
\label{figure2}
\end{figure}

\begin{figure*}
\begin{center}
\includegraphics[width=\linewidth]{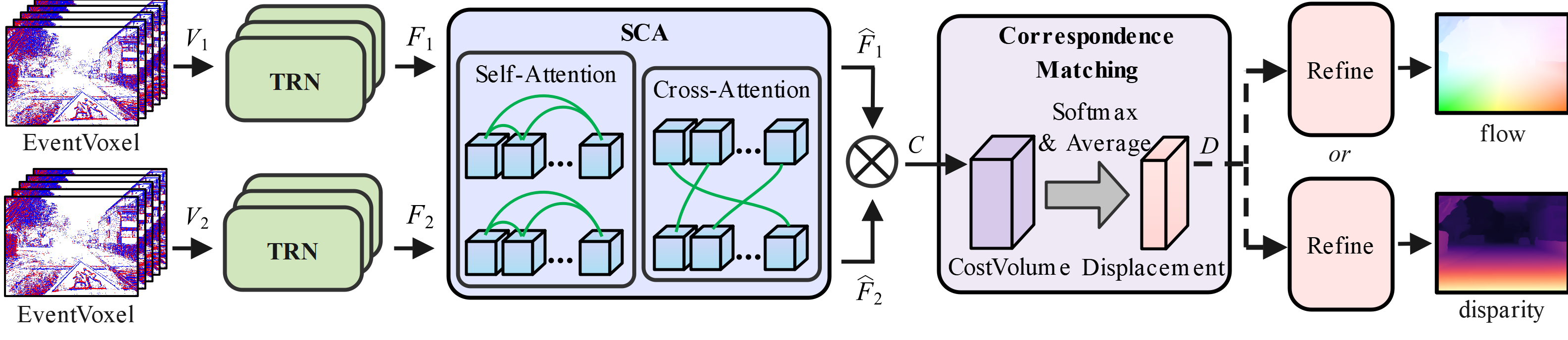}
\end{center}
\vspace{-6mm}
\caption{{Overall architecture of EMatch.} It can be divided into four parts: 1) Feature Encoding. Temporal Recurrent Network (TRN) encode reference and target event voxel $V_{1}, V_{2}$ to get initial features $F_{1}, F_{2}$. 2) Feature Enhancement. Spatial Contextual Attention (SCA) enhance features $F_{1}, F_{2}$ to obtain dense feature maps $\hat{F_{1}}, \hat{F_{2}}$ for matching. 3) Correspondence Matching. The displacement $D$ (i.e. flow or disparity) is calculated by searching for the matching features with the highest similarity between reference and target feature map $\hat{F_{1}}, \hat{F_{2}}$. 4) Refinement. The displacement $D$ are further refined to finally get flow or disparity.}
\label{figure3}
\end{figure*}

Event cameras detect changes in intensity for each pixel to generate events. Each event is represented as a tuple \((x, y, t, p)\), where $x, y$ are the pixel coordinates, $t$ is recording timestamp, and $p \in \{+1, -1\}$ is the polarity indicating an increase or decrease in brightness. During the sampling time $dt$, all recorded events can be accumulated into an event stream $E$, which can be described as:
\begin{equation}
E = \{(x_i, y_i, t_i, p_i) | i = 0,1,...,N, 0 < t_i < dt\}.
\end{equation}

Now we clarify how correspondence matching unifies event-based optical flow estimation and stereo matching, as shown in Fig.~\ref{figure2}. Let $E_1, E_2$ represent two different but related event streams as reference and target for correspondence matching. The displacement field $D=(D_x, D_y)$ describes the pixel movement from $E_1$ to $E_2$. The object function for correspondence matching both for flow and disparity can be formulated as:
\begin{equation}
\mathcal{M}(E_{1},E_{2}) = \min_{D} || \mathcal{F}(E_{1}) - \mathcal{W}(\mathcal{F}(E_{2}), D) ||,
\label{eq_match}
\end{equation}
where $\mathcal{F}(\cdot)$ denotes the feature extraction, and $\mathcal{M}(\cdot)$ is the warping operation.

In the case of optical flow estimation, the displacement field $D$ represents the motion of pixels (i.e., $f$) from start time $T_1=T$ to end time $T_2=T+\Delta T$. In the case of stereo matching, the displacement field $D$ represents the disparity (i.e., $d$) between left and right cameras. Thus, the object matching function is:
\begin{equation}
\mathcal{M}(E_{T_{1}},E_{T_{2}}) = \min_{f} || \mathcal{F}(E_{T_{1}}) - \mathcal{W}_{2d}(\mathcal{F}(E_{T_{2}}), f) ||_{2}^{2}
\label{eq1}
\end{equation}
\begin{equation}
\mathcal{M}(E_L,E_R) = \min_{d} || \mathcal{F}(E_L) - \mathcal{W}_{1d}(\mathcal{F}(E_R), d) ||_{1},
\label{eq2}
\end{equation} 
where $E_{T_1}$, $E_{T_2}$ are event streams during $(T_1-dt:T_1)$ and $(T_2-dt:T_2)$ respectively, and $E_L$, $E_R$ are event streams from left and right viewpoints respectively.

Therefore, the crux of formulation lies in extracting high-quality, discriminative features from events for dense matching. However, the sparse and asynchronous nature of events makes it challenging to obtain high-quality representations in the spatiotemporal domain. In our paper, we employ neural networks to achieve feature extraction $\mathcal{F}(\cdot)$. To obtain a synchronized input matrix while preserving the temporal continuity of the event data, we first transform the event stream $E=\{e_{i}|i=1,2,..,N\}$ into event voxel $V$ following prior works \cite{zhu2019unsupervised, eraft, TMA}:
\begin{equation}
    V(x,y,t)=\sum_{e_{i}}^{} p_{i}\cdot k_b(x-x_{i})\cdot k_b(y-y_{i})\cdot k_b(t-t^{*}_{i}),
\end{equation}
where $t^{*}_{i} = (B-1) (t_{i}-t_{0})(t_{N}-t_{0})$ and $k_b(a) = \max(0, 1-|a|)$. Next, we will demonstrate how we design our model to instantiate $\mathcal{F}(\cdot)$, and how to implement correspondence matching according to Eq.~\ref{eq_match}.

\subsection{EMatch Architecture}

\begin{figure}
\begin{center}
\includegraphics[width=\linewidth]{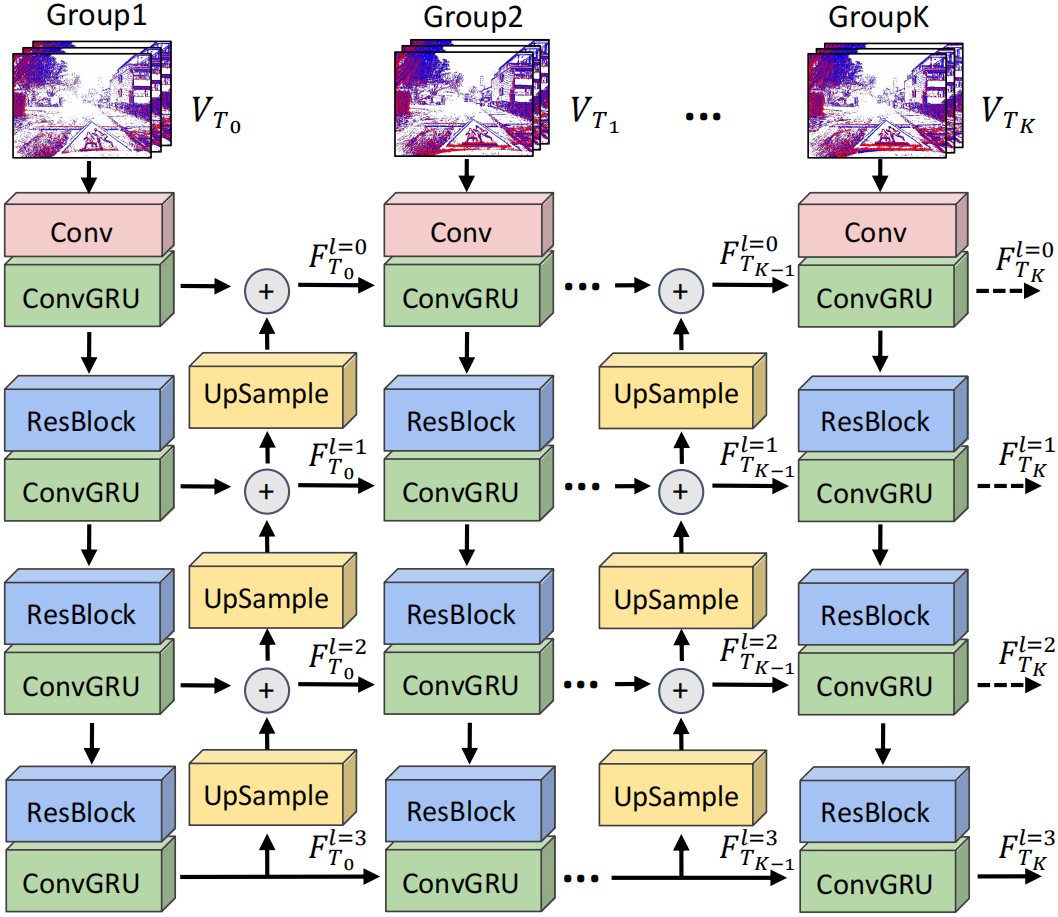}
\end{center}
\vspace{-4mm}
\caption{{Detailed architecture of TRN.} Firstly, event voxel $V$ is splited into K groups $\{V_{T_{i}}|i=0,...,K\}$ in chronological order. Then, they are fed into stacked ResBlock and ConvGRU recurrently to extract temporal features. Finally, we obtain a multi-layer features $\{F^{l=i}_{T_{K}}|i=0,1,2,3\}$ as the results. We can only use the last layer of features $F^{l=3}_{T_{K}}$ or other additional features for multi-scale optimization.}
\label{figure4}
\end{figure}

To unify event-based flow estimation and stereo matching, the correspondence matching should be applied within a shared feature domain, in which the features are task-agnostic. As shown in Fig.~\ref{figure3}, we design an event-based model named EMatch, consisting of feature encoding with Temporal Recurrent Network (TRN), feature enhancement with Spatial Contextual Attention (SCA), correspondence matching, and refinement. 
Among them, the TRN and SCA are designed to extract high-dimensional features from asynchronous and sparse events by aggregating event features temporally and spatially.

\noindent  \textbf{Temporal Recurrent Network (TRN).} Initially, we should extract features $F_{1}, F_{2} \in R^{H \times W \times  D}$ from the reference and target event voxels $V_{1},V_{2} \in R^{H \times W \times  B}$ with a parameter-sharing network. However, if we directly throw event voxels into the neural network, the temporal information preserved in the dimension of bins was wasted. Therefore, we propose to extract features from asynchronous event voxels recurrently using TRN as shown in Fig.~\ref{figure4}.

Specifically, we first split event voxel into $K$ groups according to time with $B/K$ bins for each group. For each temporal step $k$, a multi-layer recurrent network with stacked resBlock and convGRU is applied to extract current features from $V_{T_{k}}$ and aggregate history features from $F_{T_{k-1}}$ to generate current features $F_{T_{k}}$. More details can be found in the supplementary material.

Besides, TRN is capable of generating multi-scale features due to its multi-layer architecture. To further enhance information flow between layers, we introduce an additional top-to-bottom pathway, constructing a feature pyramid network for improved communication and feature integration across different scales.

\begin{figure}
\begin{center}
\includegraphics[width=\linewidth]{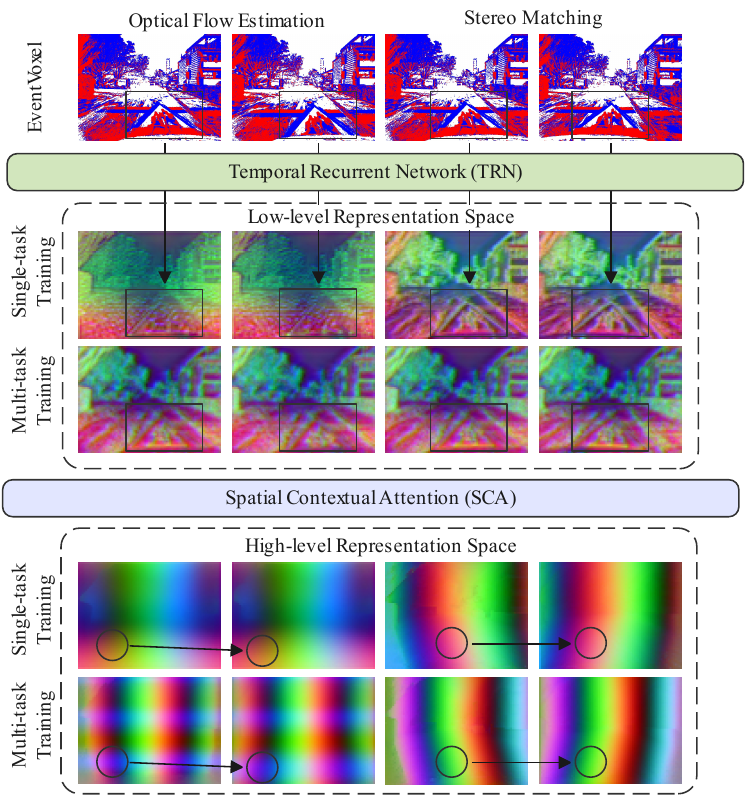}
\end{center}
\vspace{-6mm}
\caption{{Visualizing features of EMatch within different domains.} We apply PCA on intermediate features both for flow and stereo (i.e. temporal domain and spatial domain) with single-task training and multi-task training. After TRN and SCA, we can obtain a high-dimensional feature map for dense correspondence matching, and through multi-task training, the features of flow and stereo can be unified in the same representation space.}
\label{figure5}
\end{figure}

\noindent  \textbf{Spatial Contextual Attention (SCA).} Although features $F_{1}, F_{2}$ have been extracted from asynchronous event data using TRN, the sparsity of event voxels makes it less suitable for dense matching. Additional contextual information is required to assign appropriate values for pixels without triggered events. Inspired by prior works \cite{Superglue, LoFTR, gmflow, unify}, we propose to aggregate the spatial contexts globally with self-attention and cross-attention. The visualization results of this process are shown in Fig.~\ref{figure5}.

Specifically, we deploy six stacked self-attention, cross-attention, and feed-forward networks to form a feature enhancement network following previous works \cite{gmflow, unify}, and features $F_{1}, F_{2}$ are symmetrically processed to generate enhanced ones $\hat{F_{1}}, \hat{F_{2}}$. The detailed architecture of SCA can be found in the supplementary material.

\noindent \textbf{Correspondence Matching.} Given the enhanced features $\hat{F_{1}}$ and $\hat{F_{2}}$, we can compare the similarity and identify the correspondence between pixels \cite{gmflow, unify}. We first use dot product to calculate the similarity between pixels and construct the correlation volume $C_{\rm flow} \in R^{H \times W \times H \times W}$ and $C_{\rm disparity} \in R^{H \times W \times W}$.

Next, we will select the target pixel for each reference pixel with the highest similarity, constructing a correspondence to satisfy the matching function mentioned in Sec \ref{Sec3.1}. Specifically, we use softmax operation to normalize the correlation volume $C$ to obtain matching distribution $M_{\rm flow}\in R^{H \times W \times H \times W}$ and $M_{\rm disparity}\in R^{H \times W \times W}$: 
\begin{equation}
    M_{\rm flow}(a,b,x,y) =\frac{\exp[{C_{\rm flow}(a,b,x,y})]}{ {\textstyle \sum_{i,j}^{}} \exp[{C_{\rm flow}(a,b,i,j)]}}, 
\end{equation}
\begin{equation}
    M_{\rm disparity}(a,b,x) =\frac{\exp[C_{\rm disparity}(a,b,x)]}{ {\textstyle \sum_{i}^{}} \exp[C_{\rm disparity}(a,b,i)]}. 
\end{equation}

Then, the coordinate grid $G$ for the corresponding pixel can be determined with distribution $M$ by taking a weighted average of all the candidate coordinates $U_{2d} \in R^{H \times W \times 2}$ and $U_{1d} \in R^{W}$:
\begin{equation}
    G_{\rm flow} = M_{\rm flow} U_{2d} \in R^{H \times W \times 2},
\end{equation}
\begin{equation}
    G_{\rm disparity}= M_{\rm disparity} U_{1d} \in R^{H \times W}.
\end{equation}

Finally, the displacement $D$ (i.e. dense correspondence) can be obtained by computing the difference between the corresponding and initial coordinate grid. 
More details can be found in the supplementary material.

\begin{figure*}
\begin{center}
\includegraphics[width=\linewidth]{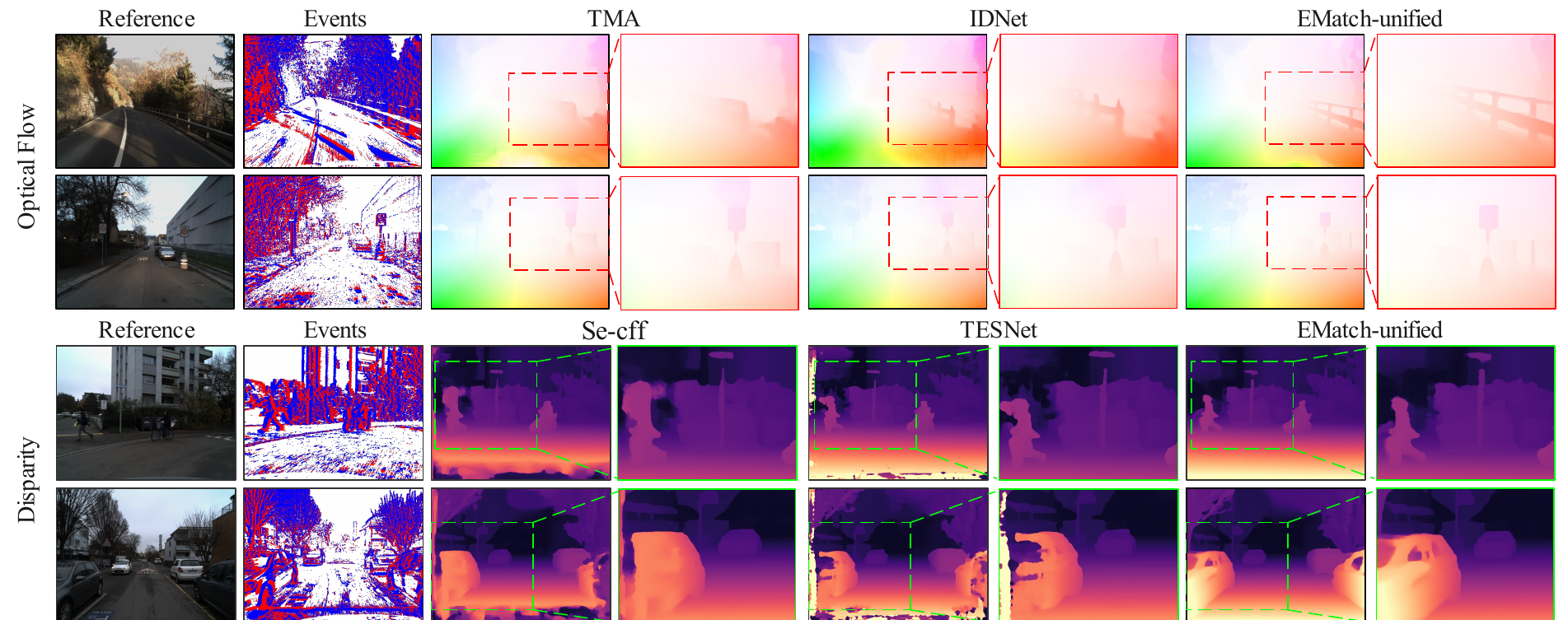}
\end{center}
\vspace{-4mm}
\caption{{Qualitative comparison of EMatch-unified with other methods.} Compared with other task-specific methods, EMatch-unified can generate results with richer details by summarizing the priors both from flow and stereo.}
\label{figure6}
\end{figure*}

\begin{table*}
    \centering
    \scalebox{0.9}{
    \begin{tabular}{c|c|c|c|c|c|c|c|c|c|c}
    \hline
    \multirow{2}{*}{Task} & \multirow{2}{*}{Method} & \multicolumn{5}{c|}{OpticalFlow} & \multicolumn{4}{c}{Disparity} \\ \cline{3-11}
    & & EPE & 1PE & 2PE & 3PE & AE & MAE & 1PE & 2PE & RMSE \\
    \hline
    \multirow{3}{*}{Flow} & E-RAFT \cite{eraft} & 0.79 & 12.74 & 4.74 & 2.68 & 2.85 & - & - & - & -\\
    & TMA \cite{TMA} & 0.74 & 10.86 & 3.97 & 2.30 & 2.68 & - & - & - & - \\
    & IDNet \cite{IDNet} & 0.72 & 10.07 & 3.50 & 2.04 & 2.72 & - & - & - & -\\
    \hline
    \multirow{3}{*}{Disparity} & DDES \cite{DDES} & - & - & - & - & -& 0.58 & 10.92 & 2.91 & 1.38 \\
    & Se-cff \cite{ConcentrationNet} & - & - & - & - & - & 0.52 & 9.58 & 2.62 & 1.23 \\
    & TESNet \cite{TESNet} & - & - & - & - & - & 0.50 & 9.20 & 2.35 & 1.18 \\
    \hline
    Flow \& Disparity & \textbf{EMatch-unified} & \textbf{0.64} & \textbf{7.86} & \textbf{2.83} & \textbf{1.69} & \textbf{2.46} & \textbf{0.47} & \textbf{8.34} & \textbf{2.17} & \textbf{1.08} \\
    \hline
    \end{tabular}}
    \caption{{Comparison of EMatch-unified with SOTA methods on DSEC benchmark.} Our unified model can simultaneously estimate flow and disparity, while maintaining SOTA performance compared with other models.}
    \label{Tab_Exp_Unifiy}
\end{table*}

\noindent  \textbf{Refinement.} In order to further eliminate the ambiguity of the matching results, we introduce a task-specific refinement head for optical flow and disparity. Specifically, we utilize a GRU iterative module similar to RAFT \cite{raft}, which generates optimized residuals based on the current optical flow query and the corresponding cost volume. The number of iterations is set to $n = 3$.

In addition, we adopt a multi-scale optimization \cite{pwcnet, FPN} to improve matching accuracy. Specifically, we construct a feature pyramid based on different resolution features (i.e., $F^{l}_{T_{K}}$ as shown in Fig.~\ref{figure4}), optimizing the matching results by warping features and estimating residual. In our paper, we adopt a two-layer feature pyramid and match pixels in global region and local region respectively. 

\subsection{Supervision}

We conduct supervised training on all outputs (including matched and refined results) with the ground truth. And we use $L_2$ loss \cite{eraft}, smooth $L_1$ loss \cite{ConcentrationNet} for optical flow estimation and stereo matching respectively:
\begin{equation}
    \mathcal{L}_{flow}=\sum_{i=1}^{N} \gamma ^{N-i} L_2(f_{gt}, f_{i}),
\label{loss_flow}
\end{equation}
\begin{equation}
    \mathcal{L}_{Disparity}=\sum_{i=1}^{N} \gamma ^{N-i} Smooth_{L_1}(d_{gt}, d_{i}),
\label{loss_disparity}
\end{equation}
where $N$ is the total number of results and $\gamma$ is a parameter to balance loss weights. In our experiments, we set $\gamma =0.8$ empirically both for flow and disparity.

\begin{table*}[htp]
\resizebox{\textwidth}{7mm}{
\small
\renewcommand\arraystretch{1} 
\begin{floatrow}
\capbtabbox{
\vspace{-2pt}
\scalebox{1}{
\begin{tabular}{c|c|c|c|c|c}
    \hline
    \multirow{2}{*}{Method} & \multicolumn{5}{c}{OpticalFlow} \\ \cline{2-6}
    & EPE & 1PE & 2PE & 3PE & AE \\
    \hline
    EV-FlowNet \cite{evflownet} & 2.32 & 55.4 & 29.8 & 18.6 & 7.90 \\
    E-RAFT \cite{eraft} & 0.79 & 12.74 & 4.74 & 2.68 & 2.85  \\
    ADMFlow \cite{ADMFlow} & 0.78 & 12.52 & 4.67 & 2.65 & 2.84 \\ 
    E-FlowFormer \cite{EFlowFormer} & 0.76 & 11.23 & 4.10 & 2.45 & 2.68 \\
    TMA \cite{TMA} & 0.74 & 10.86 & 3.97 & 2.30 & 2.68 \\
    IDNet \cite{IDNet} & 0.72 & 10.07 & 3.50 & 2.04 & 2.72 \\
    ECDDP \cite{ECDDP} & 0.70 & 8.89 & 3.20 & 1.96 & 2.58 \\
    \hline
    EMatch-single & \underline{0.67} & \underline{8.37} & \underline{3.23} & \underline{1.95} & \underline{2.52} \\ 
    \textbf{EMatch-cross} & \textbf{0.65} & \textbf{7.89} & \textbf{2.94} & \textbf{1.80} & \textbf{2.44} \\
    \hline
    \end{tabular}
}
}
{
 \caption{{Comparison for optical flow estimation on DSEC.}}
 \label{Tab_Exp_flow}
 \small
}
\capbtabbox{
\vspace{-2pt}
\scalebox{1}{
\begin{tabular}{c|c|c|c|c}
    \hline
    \multirow{2}{*}{Method} & \multicolumn{4}{c}{Disparity} \\ \cline{2-5}
     & MAE & 1PE & 2PE & RMSE \\
    \hline
    DDES \cite{DDES} & 0.58 & 10.92 & 2.91 & 1.38 \\
    E-Stereo \cite{EIStereo} & 0.53 & 9.96 & 2.65 & 1.22 \\
    DTC-PDS \cite{DTC} & 0.53 & 9.28 & 2.42 & 1.29 \\
    DTC-SPADE \cite{DTC} & 0.53 & 9.52 & 2.36 & 1.26 \\
    Se-cff \cite{ConcentrationNet} & 0.52 & 9.58 & 2.62 & 1.23 \\
    ETAM-TSCLM \cite{ETAM-TSCLM} & 0.50 & 9.20 & 2.34 & 1.14 \\
    TESNet \cite{TESNet} & 0.50 & 9.20 & 2.35 & 1.18 \\
    \hline
    EMatch-single & \underline{0.49} & \underline{8.80} & \underline{2.30} & \underline{1.10} \\
    \textbf{EMatch-cross} & \textbf{0.47} & \textbf{8.44} & \textbf{2.21} & \textbf{1.09} \\
    \hline
    \end{tabular}
}
}
{
 \caption{{Comparison for stereo matching on DSEC.}}
 \label{Tab_Exp_dis}
}
\end{floatrow}
\vspace{-2mm}
}
\end{table*}

\section{Experiments}

We evaluate our model on DSEC benchmark \cite{dsec}, a large-scale dataset that can evaluate both optical flow estimation and stereo matching. The details are as follows.

\noindent  \textbf{Implementation Details.} We implement our model in PyTorch and train it on NVIDIA RTX 3090 GPUs. We use event voxel with B=15 bins sampling from dt=100ms as inputs. During training, we utilize OneCycle learning rate scheduler and AdamW optimizer. For data augmentation, we perform spatial augmentation by randomly rescaling and stretching event voxels. And we also apply horizontal and vertical flipping with the probability of 0.5 and 0.1.

\noindent \textbf{Training Schedule.} We firstly train our model for 400k iterations with the maximal learning rate of $3 \times 10^{-4}$ and batch size of 4. The inputs are randomly cropped into the size of $288 \times 384$. Then, we further fine-tune our model for 150k iterations on the full resolution of $480 \times 640$ with the reduced learning rate of $1 \times 10^{-5}$ and batch size of 1. 

\noindent \textbf{Metrics.} For optical flow estimation, we use EPE and NPE as main metrics. For stereo matching, we adopt metrics of MAE and NPE. The end-point-error (EPE) is the $l_2$ distance between the prediction and ground truth. The mean absolute error (MAE) is the average of $l_1$ distance between the prediction and ground truth. The N-pixel error (NPE) indicates the percentage of displacement errors higher than $N$ pixels in magnitude, which is set to $N$ = 1, 2, 3.

\begin{figure}
\begin{center}
\includegraphics[width=\linewidth]{sec/Figures/EMatch-Trans.png}
\end{center}
\vspace{-4mm}
\caption{{Error curves of EMatch during single-task training and cross-task training.} Due to a better starting point, the training of EMatch-cross is significantly faster and better than EMatch-single.}
\label{figure7}
\end{figure}

\subsection{Multi-task Training} \label{Sec4.1}

We adopt multi-task training on EMatch by supervising flow and stereo alternately following Eqs.~\ref{loss_flow} and \ref{loss_disparity}. 
After training, EMatch-unified can estimate both flow and disparity simultaneously without the need for retraining on each task, which is the unique advantage of our unified model.

Due to the simultaneous supervision of flow and disparity during model training, EMatch-unified can achieve SOTA performance on both optical flow estimation and stereo matching, as shown in Table \ref{Tab_Exp_Unifiy}. 
Experimental results demonstrate that our model is able to learn more generalized feature representations from both temporal flow and spatial disparity, overcoming the limitations of finite and monotonous single-task training data.

We also provide qualitative results in Fig.~\ref{figure6} for intuitive comparison. It can be seen that training both tasks simultaneously can enhance the model's ability to perceive the world, making it more accurate in outlining object contours.

\begin{figure}
\begin{center}
\includegraphics[width=\linewidth]{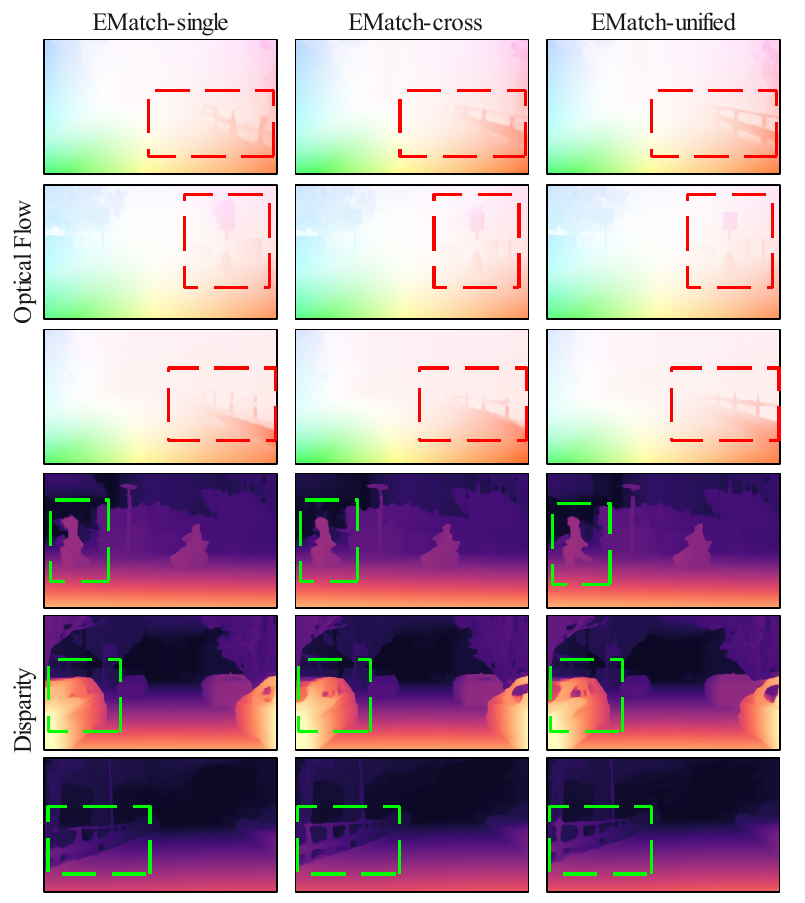}
\end{center}
\vspace{-4mm}
\caption{{Qualitative comparison for EMatch with multi-task training or single-task training.} It proves that multi-task fusion and cross-task transfer can effectively improve the effectiveness of model training, because of the communication between tasks.}
\label{figure8}
\end{figure}

\begin{table*}
    \centering
    \scalebox{0.9}{
    \begin{tabular}{l|c|c|c|c|c|c|c|c|c|c}
    \hline
    \multirow{2}{*}{Method} & \multicolumn{5}{c|}{OpticalFlow} & \multicolumn{4}{c|}{Disparity} & \multirow{2}{*}{Param} \\ \cline{2-10}
    & EPE & 1PE & 2PE & 3PE & AE & MAE & 1PE & 2PE & RMSE \\
    \hline
    Baseline & 0.94 & 15.38 & 5.98 & 3.68 & 3.13 & 0.62 & 13.25 & 3.77 & 1.34 & 2.25M \\
    \hline
    w/o TRN & 1.05 & 18.27 & 7.11 & 4.36 & 3.46 & 0.69 & 15.92 & 4.78 & 1.46 & 2.40M \\
    \quad \, \textbar -w/o convGRU \, & 0.99 & 16.70 & 6.30 & 3.89 & 3.27 & 0.66 & 14.78 & 4.35 & 1.44 & 2.01M \\
    \hline
    w/o SCA & 1.20 & 27.06 & 11.00 & 5.88 & 4.08 & 0.80 & 21.09 & 6.52 & 1.59 & 1.76M \\
    \quad \, \textbar -w/o attention \, & 1.16 & 25.80 & 9.02 & 4.84 & 4.46 & 0.71 & 16.87 & 4.98 & 1.48 & 2.05M \\
    \quad \, \textbar -w/o FFN \, & 1.00 & 17.14 & 6.49 & 3.93 & 3.30 & 0.66 & 14.73 & 4.29 & 1.41 & 1.96M \\
    \quad \, \textbar -w/o pos \, & 1.01 & 19.17 & 6.66 & 3.80 & 3.95 & 0.63 & 13.72 & 3.93 & 1.36 & 2.25M \\
    \hline
    w/o multi-scale & 1.27 & 28.76 & 11.32 & 6.38 & 4.24 & 0.88 & 25.58 & 8.02 & 1.60 &  2.31M \\
    w/o refinement & 1.56 & 25.05 & 9.41 & 5.95 & 4.64 & 0.77 & 19.51 & 5.76 & 1.56 & 0.99M \\ 
    \hline
    \end{tabular}}
    \caption{{Ablation studies for EMatch-unified.} The studies use a scaled-down model than the ones presented in Section \ref{Sec4.1} and \ref{Sec4.2}. } 
    \label{Tab_Exp_Ablation}
\end{table*}

\subsection{Cross-task Training} \label{Sec4.2}

To further demonstrate the effectiveness and flexibility of our architecture in cross-task transfer, we first adopt single-task training for optical flow estimation or stereo matching to obtain EMatch-single. Then, we adopt cross-task transfer between them to obtain EMatch-cross, during which we use the weights of the shared module from one task (such as optical flow) as the initial checkpoint and transfer it to another task (such as stereo matching) for training.

As shown in Tables \ref{Tab_Exp_flow} and \ref{Tab_Exp_dis}, cross-task trained model can achieve better results than single-task trained model, while both models achieve SOTA performance on DSEC benchmark. 
In Fig.~\ref{figure7}, we also demonstrate error curves both for EMatch-single and EMatch-cross during training, which prove that cross-task transfer can accelerate model training and make model converge better than random initialization. 

Besides, we provide a visual comparison for EMatch-single, EMatch-cross and EMatch-unified as shown in Fig.~\ref{figure8}, which intuitively demonstrates the benefits of integration between tasks for model performance.

\begin{figure}
\begin{center}
\includegraphics[width=\linewidth]{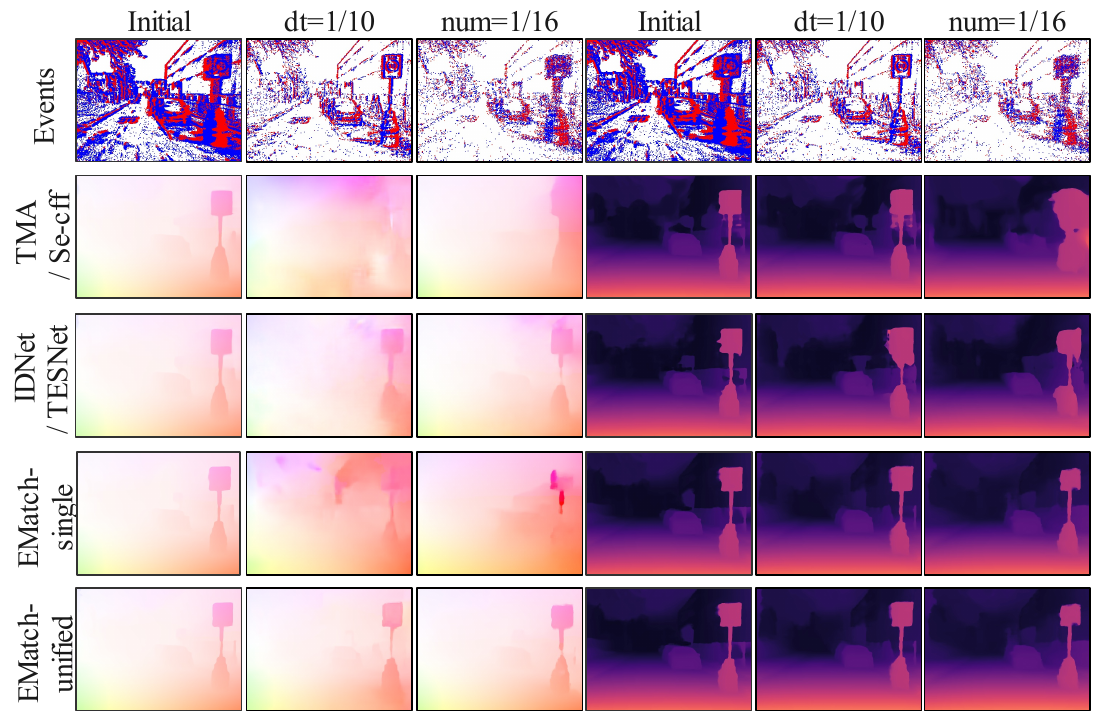}
\end{center}
\vspace{-4mm}
\caption{Comparisons for generalization performance. To simulate different data distributions, we reduce dt to 1/10 of its original setting, or keep dt unchanged but sample events at intervals to reduce its number by 1/16. Clearly, EMatch-unified has the best generalization ability.}
\label{figure9}
\end{figure}

\subsection{Ablation Studies}

We conduct a series of ablation experiments to validate the proposed modules.
To reduce computational costs, we train models for 200k iterations at the reduced resolution and fine-tune models for 10k iterations at the full resolution, both with a batch size of 1. The results are shown in Table \ref{Tab_Exp_Ablation}.
We use the scaled-down EMatch-unified with multi-task training as the baseline.

For the ablation of TRN, we use a larger CNN pyramid network \cite{gmflow} as an alternative to remove TRN (denoted as `w/o TRN'). In addition, we remove convGRU (denoted as `w/o convGRU') to cancel the recurrent iteration for TRN. The results show that the temporal operation of TRN is effective for feature extraction from event voxels.

For the ablation of SCA, we first remove the entire SCA in Baseline (denoted as `w/o SCA' ), and then we remove attention operations, feed forward networks and positional encoding separately (denoted as 'w/o attention', `w/o FFN' and `w/o pos'). It proves that the contextual feature aggregation can significantly improve final performance, while the attention operations play the greatest role.

For optimization strategies, we demonstrate the ablation results without multi-scale optimization and task-specific refinement heads (denoted as `w/o multi-scale' and `w/o refinement'). The results show that two optimization strategies are crucial for improving the model performance.

\subsection{Generalization Studies}
We test the generalization performance of our model on different event data distributions. 
We simulate a sparser data distribution by changing the original event sampling settings (reducing sampling time dt, or deleting events at intervals) to qualitatively test the performance degradation of different models.
The results are shown in Fig.~\ref{figure9}.
It can be seen that our unified model has better generalization performance compared to other single-task models, because it is trained from a wider range of data distributions (i.e. temporal flow and spatial disparity).
More results can be found in the supplementary materials.

\section{Conclusion}
We introduce EMatch, a unified framework for optical flow estimation and stereo matching, reformulating these tasks as dense correspondence matching problems within a shared representation space. By leveraging the Temporal Recurrent Network (TRN) and Spatial Contextual Attention (SCA), EMatch effectively aggregates temporal and spatial features from sparse event data. Experiments on the DSEC benchmark show that EMatch achieves state-of-the-art performance in both tasks while supporting multi-task fusion and cross-task transfer. Its unified architecture reduces deployment complexity and resource demands, making it highly suitable for real-world applications and offering new perspectives for advancing neuromorphic vision systems.   Future work will extend EMatch to additional event-based tasks and optimize it for real-time environments.

{
    \small
    \bibliographystyle{ieeenat_fullname}
    \bibliography{main}
}


\end{document}